\pdfoutput=1
\documentclass{aircc}
\usepackage{hyperref}
\usepackage{amsfonts}
\usepackage{amssymb}
\usepackage{amsmath}
\usepackage{appendix}
\usepackage{graphicx}
\usepackage{float}
\usepackage{times}
\usepackage{latexsym}

\usepackage[T1]{fontenc}
\usepackage[utf8]{inputenc}
\usepackage{pdftexcmds}
\usepackage[frozencache=true,cachedir=minted-cache]{minted} 
\usepackage{microtype}

\usepackage{inconsolata}
\title{What Makes a Good Dataset for Symbol Description Reading?}

\author{Karol Lynch$^1$, Joern Ploennigs$^{1,2}$ and Bradley Eck$^1$}
\affiliation{$^1$IBM Research Europe, Dublin, Ireland \\
            \email{\url{karol\_lynch@ie.ibm.com}},
            \email{\url{bradley.eck@ie.ibm.com}}\\
            $^2$University of Rostock, Rostock, Germany \\
            \email{\url{joern.ploennigs@uni-rostock.de}}}

\begin{document}

\maketitle

\begin{abstract}
  The usage of mathematical formulas as concise representations of a document's key ideas is common practice.  Correctly interpreting these formulas, by identifying mathematical symbols and extracting their descriptions, is an important task in document understanding.  
  This paper makes the following contributions to the mathematical identifier description reading (MIDR) task:
  (i) introduces the Math Formula Question Answering Dataset (MFQuAD) with $7508$ annotated identifier occurrences;
  (ii) describes novel variations of the noun phrase ranking approach for the MIDR task;
  (iii) reports experimental results for the SOTA noun phrase ranking approach and our novel variations of the approach, providing problem insights and a performance baseline;
  (iv) provides a position on the features that make an effective dataset for the MIDR task.
\end{abstract}


\section{Introduction}\label{sec:introduction}
% 1. Define task
The task of mathematical identifier description reading (MIDR) involves
extracting or generating descriptions of a formula's identifiers from
unstructured text like Wikipedia articles or scientific publications.
% 2. Say why this work is important!!!
Practical applications of the MIDR task include
math information retrieval~\cite{Kristianto2014,Schubotz:2017},
accessibility of mathematical content to the visually impaired~\cite{karshmer2007mathematics}, 
automated feature engineering~\cite{Lynch2019} and document understanding~\cite{lai:2022symlink}.

\begin{figure}[ht]
\centering
\includegraphics[width=0.5\linewidth]{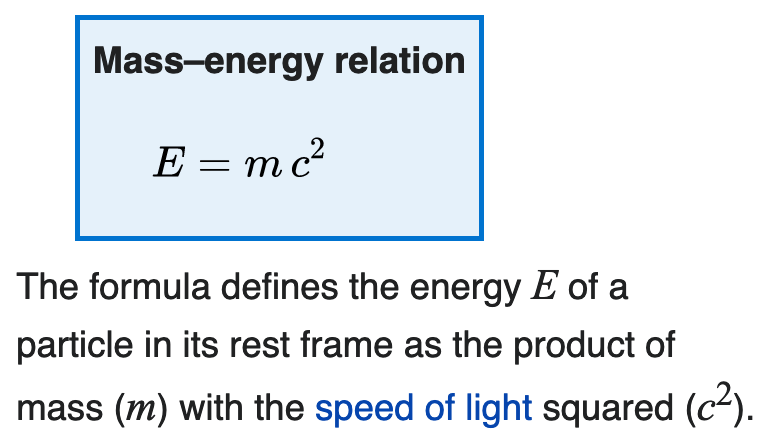}
\caption{An extract from Wikipedia showing the mass-energy equivalence relation\cite{url:meew_website}.}
\label{fig:formula_eg1}
\end{figure}

% 3. Given an example.
As an example of the MIDR task consider the mass-energy equivalence equation
and its textual context extracted from a Wikipedia article (Figure~\ref{fig:formula_eg1}).
The formula $E=m\,c^2$ includes three identifiers: $E$, $m$, and $c$.  The text below the
formula defines $E$ as \emph{the energy of a particle in its rest frame}, $m$
as \emph{mass}, and $c$ as \emph{the speed of light}.  The objective of the
MIDR task is to (automatically) extract these definitions from the article.

% 4. Why this is MIDR an interesting research problem?
%
The MIDR task can be thought of as both a form of anaphoric resolution and
reading comprehension.  Mathematical identifiers, like pronouns acquire meaning from their context.  Both identifiers and pronouns can
have several distinct meanings within a single
document~\cite{asakura:2022-building}.  Unlike pronouns, identifiers are often
explicitly defined.  However, these descriptions do not always appear as well
defined contiguous excerpts in the source document; Figure
\ref{fig:formula_eg1} is a straightforward example.  Thus extractive techniques alone are not sufficient for tackling the MIDR task.  Moreover, there are usually multiple occurrences of an
identifier, such as the constant $\pi$,  in an article~\cite{asakura:2022-building}, but its description may only be provided once or even not at all~\cite{Wolska2010}.
Recognizing identifier descriptions that either occur implicitly or are missing, poses challenges for the MIDR task.
Further, most of today's state of the art approaches for NLP tasks are based on the fine-tuning of pretrained large language models~\cite{howard:ulmfit}.  However, the performance of language models
trained solely on natural language text degrades when used with a mixture of
mathematical objects and natural language~\cite{LinWangBeyetteLiu2019}.

We believe that the availability of a relevant high quality dataset can unlock further progress in the MIDR task.
The creation of quality datasets has played a significant role in machine learning progress across many fields such as computer vision~\cite{ImageNet2009} and natural language processing~\cite{rajpurkar-etal-2018-know,marcus-etal-1993-building}.  Even in the era of foundation models and self-supervised training on a massive scale, high quality human supervision can dramatically improve model performance~\cite{Wei:FinetunedLanguageModelsAreZeroShotLearners2021}

\floatstyle{boxed}
\restylefloat{figure}

% 6. Paper Outline.
With this paper, we aim to advance the state of the art for the MIDR task by contributing 
a new dataset for validating performance on the task.  We also provide a variation of the SOTA noun phrase ranking approach. In presenting this work, we 
first give our views on the important characteristics of a dataset for training MIDR models
(Section~\ref{sec:requirements}).
A review of existing datasets and approaches appears in (Section~\ref{sec:related_work}). 
Next, we describe the development of the Math Formula Question Answering Dataset (MFQuAD), consisting of over $7500$ annotated identifier occurrences.
We then analyze MFQuAD on its own and in comparison to related data sets in Section~\ref{sec:analysis}.  Benchmark experimental results appear in Section~\ref{sec:evaluation}.  Conclusions and a discussion of future research directions are provided in
Section~\ref{sec:conclusions}.

\section{What makes a good MIDR Dataset?\label{sec:requirements}}

In this section we discuss what features make an effective dataset for the MIDR task.  Naturally the dataset should contain identifiers and their descriptions, but there are lots of different possibilities for representing them.  We argue that the following features are particularly important.

\begin{enumerate}
\item\textbf{Extracted and Generated Examples}
The dataset should contain continuous span and non-span descriptions of identifiers.
Including both reflects the reality of how documents appear in the wild.
For non-span examples it is important to provide descriptions that can be generated from the relevant source text.  We refer to these descriptions as \emph{grounded generated descriptions}.  Several such examples are described later in Section~\ref{sec:mfquad}.  The source text for grounded generative descriptions should also be included.
This supports both generative~\cite{generative2020raffel} as well as extractive techniques~\cite{devlin-etal-2019-bert}.
  
\item\textbf{Examples without Descriptions}
The dataset should contain occurrences of an identifier that are not associated with a description.
When searching for an identifier description within a document, we do not know apriori which occurrences of that identifier are associated with its definition~\cite{Wolska2010}.
Including these occurrences is analogous to how SQuAD 2.0~\cite{rajpurkar-etal-2018-know} built upon SQuAD 1.1~\cite{rajpurkar-etal-2016-squad} by adding examples where the answer to the question did not exist within the passage.

\item\textbf{Exact Identifier and Description Location}
The dataset should specify where the identifier and its corresponding descriptions are in the source article.
Otherwise,  consumers of the dataset could wrongly classify any appearance of the description text and identifier as a valid example, possibly generating invalid false positive examples.
In the case of grounded generated description examples the exact location of the source text should be provided instead.

\item\textbf{All Identifier Occurrences}
The dataset should include all occurrences of an identifier in a source article, as practical applications of the MIDR task generally involve searching for descriptions across an entire article or document.  
Further, we should not filter paragraphs based on any criteria, such as having a large amount of mathematical content or based on length.

\item\textbf{Size}
The dataset should be large enough in terms of the total number of examples in order to support learning based approaches.  The dataset should also draw its examples from a large number of source articles (as this can be considered a proxy variety for writing style and language in the dataset).
\end{enumerate}
A dataset that fulfills these requirements will support modern machine learning and NLP techniques, and reflects MIDR task use cases that occur in the wild.

\section{Related Work\label{sec:related_work}}
Accessing mathematical information in documents is a rich research area and there has been a lot of recent activity on the MIDR task.  In this section we review relevant tasks, approaches and datasets.

\subsection{Tasks and Approaches}
Tasks in mathematical information extraction focus on different levels of granularity:
(i) extracting descriptions of entire formulas~\cite{nghiem-quoc-etal-2010-mining};
(ii) extracting descriptions of mathematical expressions~\cite{Kristianto2014,LinWangBeyetteLiu2019};
(iii) extracting descriptions of mathematical identifiers~\cite{Schubotz2016a,Schubotz:2017, alexeeva-etal-2020-mathalign}.
Tasks at the identifier level of granularity include concept matching and description reading.
Concept matching attempts to match identifiers to entries in a dictionary of mathematical concepts. In description reading, the task is to describe mathematical identifiers from textual descriptions in the source document. 
We consider the fine-grained approach of the MIDR task, where we study individual elements of a formula, to be the appropriate granularity for studying a formula as it reveals a formula's important internal relationships.

Approaches to these three mathematical information extraction tasks can be categorised either as: (i) rule based approaches~\cite{nghiem-quoc-etal-2010-mining,Schubotz2016a, alexeeva-etal-2020-mathalign} or (ii) hybrid approaches that combine rules and data based techniques~\cite{Kristianto2014,Schubotz:2017,LinWangBeyetteLiu2019}.
Hybrid approaches use rules to generate candidate descriptions like noun phrases for mathematical objects and label examples to train a model to classify these as either valid descriptions or not.
Rule based approaches offer the advantages of interpretability, low data requirements and often less computational overhead.  Whereas, data based approaches to NLP tasks are dominating leaderboards~\cite{url:nlpprogress_website}, but, often come with large computational and training data requirements.
The MFQuAD dataset supports the training of hybrid (data) based approaches, but of course it can also be used for the validation of purely rules based approaches too.

\subsection{Datasets}
Recently, several datasets of interest to the MIDR task have been created.  In this section we review their contributions, covering the \emph{Gold Standard} dataset~\cite{Schubotz2016a}, the \emph{MathAlign-Eval} dataset~\cite{alexeeva-etal-2020-mathalign}, the \emph{Symlink} dataset~\cite{lai:2022symlink} and the \emph{Formula Grounding} dataset~\cite{asakura:2022-building}.

\cite{Schubotz2016a} makes many valuable contributions to the MIDR task, including the introduction of the first specialised dataset, which they named the \emph{Gold Standard}.  It contains examples drawn from 100 Wikipedia articles, and annotates 310 identifiers primarily with Wikidata IDs, but also with textual descriptions.

The MathAlign-Eval dataset introduced in \cite{alexeeva-etal-2020-mathalign}
focuses on extracting identifier descriptions from PDF documents, an important format for publications in STEM fields. It contains 757 identifiers 
from 145 formulae drawn from 116 documents from arXiv.

%In this section we present the MFQuAD dataset for the MIDR task.
% Why create a new dataset.
%The creation of the MFQuAD dataset was motivated by shortcomings of the existing datasets for the MIDR task.

Recently, \cite{lai:2022symlink} contributed the \emph{Symlink} dataset, which draws examples from the \LaTeX\ source of arXiv documents.
The \emph{Symlink} dataset frames the MIDR task as an entity and relation extraction task.
It includes three entity types: \emph{Symbol} (for symbols), \emph{Primary} (which are descriptions of a single symbol) and \emph{Ordered} (which are descriptions covering two or more symbols).
It includes four relationship types. \emph{Direct} (between a symbol and its description), \emph{Count} (relationship between a symbol and the entity that it represents the count of, e.g., in the excerpt `$k$ sets', $k$ represents the number of sets), \emph{Coref-Description} for relating two occurrences of the same description and \emph{Coref-Symbol} for annotating two occurrences of the same symbol.

% Grounded Formula Dataset
Also recently, \cite{asakura:2022-building} introduced what we refer to as the \emph{Formula Grounding} dataset.  They annotated all 12,352 occurrences of math identifiers in 15 scientific articles with their \emph{math concepts} (`instead of directly annotating each occurrence of a math token with a description, we annotated each token with a concept ID defined in the math concept dictionary').  In addition, the \emph{Formula Grounding} dataset contains $938$ `text spans that can be used as bases for a human to ground formula tokens'.  These `sources of grounding' may or may not be standalone valid descriptions themselves.
The focus on concepts, rather than textual descriptions allowed the authors to study the task of coreference resolution for mathematical identifiers (e.g., do two occurrences of a symbol $x$ represent the same entity?).

% Conclusion
Whilst all these datasets made significant contributions, none provide all the features discussed in Section~\ref{sec:requirements}.  A detailed analysis of how these datasets satisfy the features described in Section~\ref{sec:requirements} is presented in Section~\ref{sec:analysis}.

\section{The MFQuAD Dataset\label{sec:mfquad}}
% Introduce MFQuAD
We create a new dataset for MIDR task that includes all the important features described in Section~\ref{sec:requirements}.
We address the question of what articles and formulas to annotate by using the same set of identifiers and formulas as an existing dataset.
We chose the~\emph{Gold Standard} dataset as the other existing datasets are either PDF based~\cite{alexeeva-etal-2020-mathalign} (which introduces a vision component that is outside the scope of the textual MIDR task), or are not generally available~\cite{asakura:2022-building}.

Also, as we wish to support modern NLP techniques such as reading comprehension we model the examples in our dataset on the SQuAD 2.0 dataset~\cite{rajpurkar-etal-2018-know} (where the SQuAD dataset contains answers, we include identifier descriptions).  Thus we name it the \emph{Math Formula Question Answering Dataset} (MFQuAD).

The remainder of this section is organised as follows:  Section~\ref{sec:mfquad_creation} describes how we created the MFQuAD dataset, Section~\ref{sec:example-representation} describes the data fields that we provided for each example, and finally Section~\ref{sec:mfquad-study} provides a qualitative and quantitative study of the MFQuAD dataset.

\subsection{MFQuAD Creation\label{sec:mfquad_creation}}
MFQuAD includes all 310 identifiers and 100 formulas/expressions of the \emph{Gold Dataset}, which were drawn from 100 Wikipedia articles.
In Section~\ref{sec:example-preparation} we outline how examples were preprocessed and prepared for annotation and in Section~\ref{sec:dataset-annotation} we describe the annotation process.

\subsubsection{Example Preprocessing\label{sec:example-preparation}}
We create an annotated entry for each occurrence of an identifier in its source article with 7508 examples in total.
Thus MFQuAD's scale is in line with the larger datasets described in Section~\ref{sec:related_work}~\cite{lai:2022symlink, asakura:2022-building}.
The chosen representation of the Wikipedia article excerpts is their raw Wikitext format, in which formulas are represented in a \LaTeX\ dialect.  This choice supports experimenting with different approaches to formula processing.
Identifier occurrences were retrieved using a comprehensive regular expression whose goal was to match all identifier representations (including HTML, Unicode, and \LaTeX\ formats).
Each entry includes its identifier occurrence's context and all the descriptions, if any, for the identifier in the context.  The start and end positions of all descriptions are also noted.
As a preprocessing step, inline references are stripped from the raw Wikitext as they don't appear inline in the presentation format of a Wikipedia article.

Another decision involved selecting how much of an identifier's context to present to annotators.  If too little context is presented, descriptions may be missed, but if excessive context is presented, it increases the overhead of the annotation process.  We took a conservative approach, selecting 200 characters from before and 200 characters after the occurrence of the identifier, which is then extended to a paragraph boundary.  Thus passages are at least one paragraph and at least 400 characters (plus identifier length) in length.
 This was then manually annotated by by adding all descriptions for that identifier that are defined within the extracted context.

\subsubsection{Example Annotation\label{sec:dataset-annotation}}
We annotate two types of identifier descriptions in MFQuAD, textual descriptions and Wikilink descriptions (which are references to other pages within Wikipedia).
In practice Wikilinks are mapped to surface text descriptions anyway, so the number of identifiers with descriptions is unchanged.
For completeness we included numerical descriptions when annotating MFQuAD. But as these are all example values for variables in MFQuAD, rather than constants such as the speed of light, they were not treated as valid descriptions during evaluation.
Similar to previous authors~\cite{Kristianto2014,alexeeva-etal-2020-mathalign} we only consider definitions, not properties, as valid descriptions, even if the property is mathematically equivalent to the definition.
In the case of non-span identifier descriptions, we include both the source evidence and the actual derived description (i.e., grounded generated descriptions).
We also allow descriptions that are sub-strings of other longer descriptions, if and only if both can be considered as valid and complete.
For example, we consider both `the density of water (units of mass per volume)' and `the density of water' as valid for the identifier $\rho$ from the Wikipedia article on \emph{Darcy's Law}.
Annotation was performed by a senior researcher with a background in discrete mathematics and who is a native English speaker.

\subsection{MFQuAD Example Representation\label{sec:example-representation}}
We now describe the JSON object that is created for each identifier occurrence.
Consider Listing~\ref{raw-passage-example}, which shows an entry for the identifier $T_c$ from the `Modigliani–Miller\_theorem' Wikipedia article (specifically the version in the \emph{gold standard dataset}).
\begin{listing}
    \begin{minted}[frame=single,
                   framesep=1mm,
                   linenos=false,
                   xleftmargin=8pt,
                   tabsize=1,
                   breaklines,
                   breaksymbolleft=]{js}
{
"occurrence": 1,
"sanitised": true,
"id_mini_context": "'\n* <math>T_c</math> ''",
"matched_id_representation": "T_c",
"id_article_offset": 6114,
"id_passage_offset": 512,
"between_math_tags": true,
"same_as_previous": false,
"passage": "\n===Proposition II===\n:<math>r_E = r_0 + \frac{D}{E}(r_0 - r_D)(1-T_C)</math>\n\nwhere:\n\n* <math>r_E</math> ''is the required rate of return on equity, or cost of levered equity = unlevered equity + financing premium.''\n* <math>r_0</math> ''is the company cost of equity capital with no leverage (unlevered cost of equity, or return on assets with D/E = 0).''\n* <math>r_D</math> ''is the required rate of return on borrowings, or [[cost of debt]].''\n* <math>{D}/{E}</math> ''is the debt-to-equity ratio.''\n* <math>T_c</math> ''is the tax rate.''\n\nThe same relationship as earlier described stating that the cost of equity rises with leverage, because the risk to equity rises, still holds. The formula, however, has implications for the difference with the [[Weighted average cost of capital|WACC]]. Their second attempt on capital structure included taxes has identified that as the level of gearing increases by replacing equity with cheap debt the level of the WACC drops and an optimal capital structure does indeed exist at a point where debt is 100%.",
"answers": [
  {
    "wikilink": false,
    "explicit_answer": true,
    "raw_span": "the tax rate",
    "surface_span": "the tax rate",
    "deduced_answer": "the tax rate",
    "references_this_id_occurrence": true,
    "start_pos": 528,
    "end_pos": 540
  }
],
"digest": "7de9be07074e547e541b69ef0a55125a"
}
    \end{minted}
    \caption{MFQuAD Dataset Example} 
    \label{raw-passage-example}
\end{listing}

\begin{itemize}
    \item The \emph{occurrence} field specifies the occurrence order for this entry (e.g., Listing~\ref{raw-passage-example} shows the first occurrence of the identifier $T_c$ in the article ).
    \item 
    The \emph{sanitised} field indicates whether or not the raw passage width was adjusted, in order to avoid splitting a Wikitext template or element.
    \item
    The \emph{id\_mini\_context} field shows 10 characters either side of the identifier.
    \item
    The \emph{matched\_id\_representation} field shows which representation of the identifier was matched.
    \item
    The \emph{id\_article\_offset} field specifies the character offset for the identifier in the entire article.
    \item
    The \emph{id\_passage\_offset} field specifies the character offset from the beginning of the passage.
    \item
    The \emph{between\_math\_tags} field specifier whether or not the identifier occurrence was between $\langle math \rangle$ tags.
    \item
    The \emph{same\_as\_previous} field specifies whether or not the passage in this occurrence is identical to that of the previous occurrence.
    \item The \emph{passage} field contains the context extracted for this occurrence.
    \item The \emph{answers} field is an array of answer objects, including all descriptions for the identifier in this passage. An answer object contains the following elements.
\begin{itemize}
    \item The \emph{wikilink} field indicates whether or not this description is a Wikilink.
    \item The \emph{explicit\_answer} indicates whether or not this is an exact contiguous description (i.e., a span) defined using natural language (as opposed to in a table or mathematical notation).
    \item The \emph{raw\_span} field is a subspan of the raw Wikitext representing an identifier description (in the case of implicit descriptions this might be the answer source rather than an exact specific description).
    \item The \emph{surface\_span} field is a subspan of the surface or presentation format representing an identifier description (in the case of implicit descriptions this might be the answer source rather than an exact specific description).
    \item The \emph{deduced\_answer} field specifies the precise description, whether or not it appears as a span in the source.
    \item The \emph{references\_this\_id\_occurrence} field specifiers whether or not this identifier description was defined in reference to this specific occurrence (as there can be multiple occurrences with a single passage).
    \item The \emph{start\_pos} field specifies the passage offset to the first character of the description.
    \item The \emph{end\_pos} field specifies the passage offset to (one position after) the last character of the description.
    \item The \emph{digest} field is a hash of the passage field.
\end{itemize}
\end{itemize}

\subsection{MFQuAD Study\label{sec:mfquad-study}}
In this section we present statistics of the MFQuAD dataset that are relevant to the MIDR task in Section~\ref{sec:dataset-statistics}, followed by a breakdown of different classes of descriptions in Section~\ref{sec:description-analysis}.

\subsubsection{Example Statistics\label{sec:dataset-statistics}}
To inform  discussion of MFQuAD and the MIDR task, we analyzed some statistics of MFQuAD. 
MFQuAD consists of 100 formulas taken from 100 Wikipedia articles, with a total of 310 identifiers, and 7508 annotated identifier occurrences.

\begin{figure}[htbp]
\centering
\includegraphics[width=0.7\linewidth]{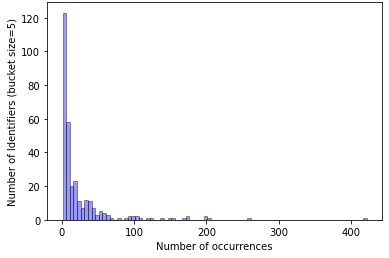}
\caption{Distribution of identifier occurrence counts (i.e. the number of times an identifier occurs or is mentioned in an article) in the MFQuAD dataset.}
\label{fig:id_occ_dist}
\end{figure}

Figure~\ref{fig:id_occ_dist} shows the source article occurrence counts for the identifiers of MFQuAD.  The counts follow a Pareto distribution as observed in \cite{Schubotz2016a}, as most identifiers only occur a few times, with at most a handful occurring many times.  The minimum and maximum occurrence counts are $1$ and $422$, respectively. The mean and median occurrence counts are $24.2$ and $9$, respectively.  High identifier occurrence counts are a challenging aspect of the MIDR task, as they increase the amount of text that must be searched for an identifier description.

\floatstyle{plain}
\restylefloat{figure}

\begin{figure}[hbtp]
\centering
\includegraphics[width=0.7\linewidth]{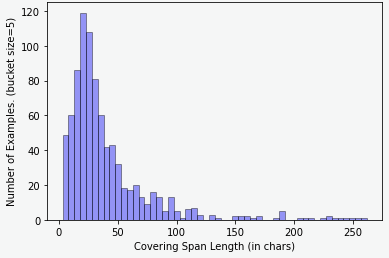}
\caption{Distribution of the lengths (in characters) of the minimum covering spans of identifiers and their descriptions in the MFQuAD dataset.}
\label{fig:min_covering_span_dist}
\end{figure}

The length of the minimum span containing both an identifier and its associated description, informs us how much context is needed to extract identifier descriptions.
Figure~\ref{fig:min_covering_span_dist} shows the distribution of the minimum covering span lengths in characters of all examples in the MFQuAD dataset.
% Ohoh, tables
Identifiers that are defined using a table's row/column structure (which occurred in a handful of examples), are omitted as they have a different distribution.
The mean, median, and maximum covering span lengths are are $40$, $28$, and $262$ characters, respectively.  The $97.5\,\%$ quantile is $164.3$. Thus including $165$ characters on both sides of an identifier (330 characters in total) would capture $97.5\,\%$ of identifier descriptions in the MFQuAD dataset.

\floatstyle{plain}
\restylefloat{figure}

\begin{figure}[hbtp]
\centering
\includegraphics[width=0.7\linewidth]{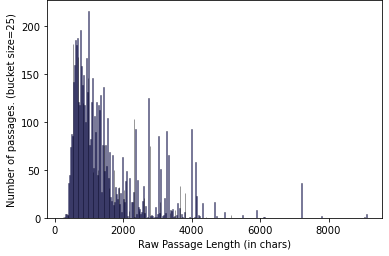}
\caption{Distribution of the identifier context lengths (i.e., the passage lengths) in characters of the MFQuAD dataset.}
\label{fig:passage_lengths}
\end{figure}

The length of identifier contexts is important as they should be long enough to contain all associated descriptions.
The minimum and maximum context lengths in the MFQuAD dataset are 231 and 9119 characters, respectively.
The mean and median context lengths are $1454.92$ and $1089$, respectively.
Figure~\ref{fig:passage_lengths} shows the distribution of the identifier context lengths (i.e., the passage lengths) in characters of the MFQuAD dataset.
It's clear that the context lengths used in MFQuAD are significantly longer than its minimum covering span lengths.

\floatstyle{plain}
\restylefloat{figure}

\subsubsection{Description Analysis\label{sec:description-analysis}}
The source articles contain (textual) descriptions for 224 out of the 310 identifiers (around 72.3\,\%).  This is consistent with the findings in \cite{Wolska2010} that around 30\,\% of the symbolic elements in scientific papers are not explicitly described in the text (at least not within their first $5$ occurrences).
We classify identifier descriptions as either \emph{implicit} or \emph{explicit}. Explicit examples have a contiguous exact description (i.\,e.~a span) defined using natural language as opposed to in a table or mathematical notation. Examples that are not explicit are classified as implicit.
Out of the 224 identifiers with descriptions in MFQuAD, 175 have at least one explicit description.
The remaining 49 identifiers have only implicit descriptions, and among these we identify the common patterns that give rise to implicit examples:

\begin{itemize}
    \item \textbf{Non-span Examples:} $38$ of these $49$ identifiers (i.\,e., $77.6\,\%$) have descriptions which do not occur as a span (i.e., contiguous tokens) within the source article.  These are examples of grounded generated descriptions.  The most common pattern for non-span descriptions (applies to $25$ out of the $38$ identifiers with non-span descriptions) is when multiple identifiers are defined together, so that an article contains a plural, rather than a singular.  For example, the extract ``$\beta$ and $I$ are parameters of the model'', from the source article \emph{331 model}, which describes the identifiers $\beta$ and $I$ together.  But a description for $I$ alone, which we want, is \emph{a parameter of the model}, but that is not contained in the source article.
    Another common pattern of non-span descriptions (applies to $9$ out of the $38$ identifiers with non-span descriptions), occurs when an identifier is used to denote a count.  For example, the extract ``with k factors'', from the source article \emph{Binomial theorem}, which defines k as the number of factors, but this is not a description that can be extracted as a span from the source.
    Another pattern of non-span descriptions (applies to $2$ identifiers) is indirect definitions, (consider the excerpt ``a polynomial in x and y'', from the source article \emph{Binomial\_theorem}, from which we can deduce that x is a variable of a polynomial even though the term \emph{variable} is not explicitly used).
\item
\textbf{Mathematical Notation:} $7$ of the $49$ identifiers (i.\,e., $14.3\,\%$) are defined using mathematical notation.  For example consider $y \in \mathbb{R}^b$ which defines y as an n-tuple of b real numbers, but it both requires understanding mathematical notation as well as the convention that $\mathbb{R}$ represents real numbers.  Further consider $e_1 \triangleq z_1 - u_x(\mathbf{x})$ which requires understanding of the mathematical notation $\triangleq$, and also doesn't provide a textual description for $e_1$.
\item
    \textbf{Table:} $4$ of the $49$ identifiers (i.\,e., $8.2\,\%$) are defined via a table.
\item
    \textbf{Symbol Decomposition:} $1$ of the $49$ identifiers, $H$, from the source article \emph{Analytic capacity} requires, the decomposition of the symbol $H^1$, which is defined as the `1-dimensional Hausdorff measure', to deduce that the definition of $H$ is the \emph{Hausdorff measure}.
\end{itemize}
For reference, a complete list of implicit examples is provided in Appendix~\ref{sec:appendix-implicit-egs}.
\section{Analysis\label{sec:analysis}}
In this section we analyse MFQuAD and the four datasets introduced in Section~\ref{sec:related_work}, across the dimensions introduced in Section~\ref{sec:requirements}. Table~\ref{tab:midr_dataset_comparison} allows for easy comparison across these criteria with MFQuAD.

Aspects of the \emph{Gold Standard} dataset which are at odds with the features outlined in Section~\ref{sec:requirements} include:
(i) that it is not self-contained as descriptions are taken from external sources (i.\,e.\ not directly from the article that the source formula appears in);
(ii) majority of descriptions are Wikidata Identifiers as opposed to textual descriptions used in the MIDR task;
(iii) it does not specify where in an article a description is extracted from or what identifier occurrence it refers to.
In \cite{Schubotz:2017} it is advised against using this dataset for training, due to ``the consultation of tertiary sources by the domain experts who created the gold standard''.

The \emph{MathAlign-Eval} dataset~\cite{alexeeva-etal-2020-mathalign} was annotated after a filtering process was applied to its source articles. Specifically, papers ``that contained more than 5 formulae'', are omitted.  Further, from the remaining papers, formulas ``whose context had more than 50 characters contained within math environments'' (context is defined as the 3 paragraphs before and after a formula here) are excluded.
For comparison, consider that the median identifier context in MFQuAD contains 228 non-whitespace characters in math environments, and 80\,\% of contexts contain 80 or more characters.  Applying just the latter restriction to MFQuAD would therefore exclude over 80\,\% of all examples.
As a consequence, an evaluation performed on the \emph{MathAlign-Eval} dataset may not by representative of the general performance for the MIDR task.

%Issues with Symval
The \emph{Symlink} dataset~\cite{lai:2022symlink} does not include generative examples or examples for identifier occurrences that are not associated with a description (only correferenced symbols are considered).  Furthermore, short paragraphs are not considered in this dataset.

Finally, the Formula Grounding dataset~\cite{asakura:2022-building} does not include generative, extractive or NULL examples, due its focus on the coreference resolution task which is more suited to concept mappings.

\begin{table*}[t]
  \centering
  \resizebox{\textwidth}{!}{%
  \begin{tabular}
      {
      | p{1.7cm}    |       p{.8cm}   |   p{1.6cm}    |      p{1.6cm}      |        p{1.6cm}     |            p{1.6cm}     | p{1.4cm}   |  p{1.4cm} | p{1cm} |}
      \hline
    Dataset             &  \#Docs & \#Labelled Examples   & Generative Examples   & Extractive Examples & NULL Examples  & Unfiltered     & Exact Location & Whole Doc\\
    \hline
    Symlink             & 101         & 7108              & No                      & Yes                & No             & No              & Yes & No\\
    \hline
    Formula Grounding   & 15          & 12352             & No                      & No                 & No            & Yes           & Yes & Yes\\
    \hline
    MathAlign           & 116         & 990               & No                      & Yes                & No             & No                                & Yes & No\\
    \hline
    Gold Standard        & 100         & 337               & No                      & No                 & No             & Yes                               & No & Yes\\
    \hline
    MFQuAD              & 100         & 7508              & Yes                     & Yes                & Yes            & Yes                         & Yes & Yes\\
    \hline
  \end{tabular}}
  \caption{This table compares four existing datasets~\cite{lai:2022symlink, asakura:2022-building, alexeeva-etal-2020-mathalign, Schubotz:2017} for the MIDR task and our \emph{MFQuAD} dataset across the features discussed in Section~\ref{sec:requirements}.}
  \label{tab:midr_dataset_comparison}
\end{table*}
% Notes: 1. Symlink Labelled Examples = Primary + Ordered
%        2. Gold Labelled Examples 310 + 27 : 310 = Num identifiers. 27 Number with two descriptions.
%        3. MathAlign Labelled Examples  = Total Number of Identifier Occurrences

In conclusion, as shown in Table~\ref{tab:midr_dataset_comparison}, MFQuAD is the only dataset satisfying all the requirements for an effective MIDR dataset - all other referenced datasets lack three or more criteria.  Further, MFQuAD's ease of use and format similarity to popular NLP dataset SQuAD 2.0, supports widespread adoption.

\section{Evaluation\label{sec:evaluation}}
In order to validate the fitness of the the MFQuAD dataset for the MIDR task we conduct experiments using the state-of-the-art noun phrase ranking (NPR) approach of \cite{Schubotz:2017}.
Furthermore, we study several variations of the NPR approach, which we describe in Section~\ref{sec:approach}.  We describe the experiments in Section~\ref{sec:experiments} and discuss the results in Section~\ref{sec:results}.
\subsection{NPR Approach\label{sec:approach}}
In Section~\ref{eval-general-preprocessing} we outline the main mathematical preprocessing steps, in Section~\ref{sec:npr-algo} we outline the main steps in the NPR approach and in Section~\ref{npr-feature-set} we summarise the set of features that the NPR classification models are trained on.  Technical implementation details are provided for reference in Section~\ref{sec:npr-technical}.
In addition to the approach in \cite{Schubotz:2017} we evaluate several variations based on the use of distributed word representations \cite{Mikolov2013}, dimensionality reduction using PCA, an alternative candidate generation strategy and other modelling algorithms (beyond SVM which is used in \cite{Schubotz:2017}).

\subsubsection{Mathematical Preprocessing\label{eval-general-preprocessing}}
In preprocessing for the MIDR task mathematical objects naturally receive special attention.
Wikipedia articles are created in a rich format called Wikitext markup, from which we extract a surface text representation prior to modelling.  
In Wikitext mathematical formulas and expressions are mostly created using a variant of \LaTeX~called \emph{texvc}, which we process using Wikipedia's own math processing tool~\cite{url:mathoid_webpage}.
Based on the approach in \cite{Schubotz:2017}, we preprocess the identifier context by replacing mathematical expressions and formulas with placeholders.  Expressions are replaced with \emph{expression}$N$ and formulas with \emph{formula}$N$, where $N$ denotes its occurrence order in its source article.
Finally, we normalise different encodings of the same mathematical identifier (e.\,g. HTML, Unicode, or \LaTeX\cite{url:wikipedia_formula}) with a consistent surface representation.  Note that although we reproduce the mathematical preprocessing approach of \cite{Schubotz:2017} here, the MFQuAD dataset supports and we encourage the exploration of other forms of formula preprocessing.

\subsubsection{NPR Algorithm\label{sec:npr-algo}}
The first step in the NPR algorithm is to extract all candidate descriptions from sentences in the source article that contain the target identifier.
Two different approaches for candidate description generation are considered: the noun-(preposition)-noun or adjective-noun phrases approach (NNAN approach) used in \cite{Schubotz:2017} and an approach we introduce using base noun phrases created from noun chunks extracted by the spaCy software library\cite{url:spacy_website}.
For the purposes of training, we first normalise both candidate and actual descriptions by removing articles, punctuation, converting to lower case, and normalising white space.  A candidate is labelled as positive iff it is a substring of an actual description. Otherwise it is labelled negative. Note that the scoring strategy is described in Section~\ref{sec:experiments}.

The NNAN candidate generation strategy results in 1307 positive examples and 9976 negative examples (a ratio of 1 to 7.63) on the MFQuAD dataset, whilst the noun chunk (NC) strategy results in 1067 positive examples and 10824 negative examples (a ratio of 1 to 10.14).
As there are significantly more negative than positive examples, prior to training the number of negative examples is downsampled~\cite{Schubotz:2017} using different sampling ratios.
The features described in Section~\ref{npr-feature-set} are then generated, and standardised by subtracting the mean and scaling to unit variance.
As a configuration option, before modelling, we reduce the dimensionality of the feature set by applying PCA.
Finally, in addition to the SVM model used in \cite{Schubotz:2017}, we also train with gradient boosting, random forests and an ensemble of all the above using a soft voting classifier.

%%%%%%%%%%%%%%%%%%%%%%%%%%%%%%%%%%%%%%%%%%%%%%%%%%%%%%%%
\subsubsection{NPR Feature Set\label{npr-feature-set}}
We now provide a high level overview of the features used in the NPR approach, with a detailed description left to Appendix~\ref{app:appendix-npr-features}.
Since we reproduce the NPR approach of~\cite{Schubotz:2017}, our feature set is a superset of the feature set of \cite{Schubotz:2017} (for configurations reproducing the SOTA approach verbatim we disable the additional features).

The features of \cite{Schubotz:2017} were gathered from three previous approaches~\cite{Kristianto2014,Schubotz2016a,Pagel2014} and can be grouped into the following categories: (i) Pattern Matching, (ii) Information Retrieval, (iii) Text and (iv) Dependency Graph.
The pattern matching features are taken from \cite{Pagel2014} (e.\,g.~``\emph{<id> (is|are) <candidate>}'') and \cite{Kristianto2014} (e.\,g.~``\emph{is there a colon between the identifier and candidate?''}).
From the same source we include a superset of the text features (e.\,g.~``\emph{two tokens from before and after the <candidate>}'').  Text features are represented as a matrix of token counts and/or a matrix of unigram, bigram and trigram counts.
We include a superset of the dependency graph features (e.\,g.~``\emph{the first three tokens on shortest path from <candidate> to <identifier>}'') from \cite{Kristianto2014}.
We include the information retrieval style features (e.\,g.~``\emph{number of tokens distance between an <identifier> and <candidate>}'') from \cite{Pagel2014}.

Finally we encode the following features using a distributed representation of text, which were not used in previous NPR approaches:
(i) two tokens from before and after candidate; %get\_definien\_context\_embedding,
(ii) three tokens from before and after identifier; % get\_id\_context\_embedding
(iii) the first verb between candidate and identifier; % get\_first\_verb\_embedding,
(iv) the text between the candidate and the identifier. %get\_link\_embedding

\subsubsection{NPR Technical Details\label{sec:npr-technical}}
We now provide the technical details of our implementation of the NPR approach.
The NPR approach is dependent on some core NLP capabilities for multiple tasks involved in both candidate and feature generation.
For candidate description generation these include sentence segmentation, POS Tagging and base noun phrase extraction.
For feature generation these include dependency parsing, distributed representations and other bag-of-word style textual feature representations.
The NPR approach is also dependent on modelling capabilities for building a classifier.
The NPR approach~\cite{Schubotz:2017} is implemented in Java using Stanford CoreNLP\cite{url:corenlp_website} for NLP functionality and \emph{Weka}\cite{url:weka_website} for modelling and some feature generation.
The past half decade there has seen immense progress in NLP capabilities with several state-of-the-art NLP libraries now available in Python, such as spaCy~\cite{url:spacy_website} and HuggingFace Transformers~\cite{url:transformers_website}.
Due to its ease of use, and powerful NLP and machine learning libraries, our NPR approach is implemented in Python using spaCy for NLP capabilities and scikit-learn~\cite{url:scikit_website} for machine learning.
Specifically we used version 3.1 of both the spaCy library and its \emph{en\_core\_web\_lg} model for all NLP functionality in the NPR approach.
Evaluation showing the effectiveness of the \emph{en\_core\_web\_lg} model for the sentence segmentation, POS (Part of Speech) tagging and dependency parsing tasks is available on the spaCy website~\cite{url:spacy_website}.
The \emph{scikit-learn} library provided us with implementations of gradient boosting, random forests, SVMs, PCA and bag-of-word style textual feature representations.

\subsection{Experiments\label{sec:experiments}}
We run experiments for $240$ configurations of the NPR approach (48 different feature combinations times 5 different models) based on the description in Section~\ref{sec:approach}.  These evaluated configurations include a reproduction of the SOTA approach~\cite{Schubotz:2017}, as well as other variations of the NPR approach.
We evaluated both the base noun phrase and the NNAN candidate description generation approaches.
The resulting pool of candidate descriptions is imbalanced and we downsample the negative examples by taking 1, 2, or 4 times the number of positive examples.
As there are many features, we evaluate the approaches with and without PCA for feature compression with $50$, $100$, and $200$ components.
We also assess if the addition of word vector based features improves the results.
In addition to the SVM configuration used in \cite{Schubotz:2017} (with hyperparameters $\gamma=0.0186$,  kernel=`rbf',  C=1.0) we appraise several other modelling options.
These include SVM with $\gamma=n_{features}^{-1}$, kernel=`rbf', C=1.0 (SVM\textsubscript{def}), gradient boosting (GBM) with 100 estimators and max depth of 3, random forest (RF) classifier with 100 estimators and max depth of 3) and a soft voting classifier (VC) which combines the GBM, RF and SVM classifiers.
We use 5-fold cross-validation with consistent splits used for all approaches for the evaluation of models that we train on the MFQuAD dataset.  We report the metrics Exact Match (EM) and F1-score as used in the official SQuAD scoring script~\cite{rajpurkar-etal-2018-know}.

Non-span descriptions require special consideration during training, as they are not contiguous sub-strings in the source article.  In these cases, we use the ``surface\_answer'' field (see Section~\ref{sec:mfquad}) for labelling during training, as we are using an extractive approach.  The actual ``deduced\_answer'' field is used for scoring.

As well as reporting results across all $310$ identifiers, we breakdown the results for some interesting identifier groupings.
\begin{itemize}
    \item \textbf{All Identifiers} - all $310$ identifiers in the MFQuAD dataset.
    \item \textbf{Explicit Descriptions} - the $175$ identifiers with at least one explicit descriptions.
    \item \textbf{Implicit Descriptions} - the $49$ identifiers that only have implicit descriptions.
    \item \textbf{Any Descriptions} - the $224$ identifiers that have a description.  This is the union of the \emph{Explicit Descriptions} and the \emph{Implicit Descriptions} classes (i.e., $175+49$ identifiers).
\end{itemize}
Results for groupings where all identifiers have a description (i.e., the \emph{Explicit Descriptions}, \emph{Implicit Descriptions} and \emph{Any Descriptions} groupings) are based on the top non-null prediction.
Results for identifiers in \emph{All Identifiers} are based on the top prediction, regardless of whether it is null or non-null.  This is because identifiers in  \emph{All Identifiers} may or may not be associated with a description.
\subsection{Results Discussion\label{sec:results}}
We now discuss the results from the experiments run over $240$ configurations of the NPR approach (48 different feature combinations times 5 different models).
To keep the report concise we don't include the results for all $240$ configurations in Table~\ref{tab:npr-results}, instead we report the top 3 results for each identifier category.
Recall that the SOTA approach of \cite{Schubotz:2017} consists of an SVM model using the NNAN candidate description generation strategy, without the use of dimensionality reduction or distributed word vectors.

A gradient boosting based modelling variation of the NPR approach produced the best results across all identifiers (i.e. identifiers with and without descriptions) with an exact match score of $0.477$ and an F1-Score of $0.529$ (c.f., Table~\ref{tab:npr-results}).
Across all configurations in our evaluation an identifier is considered to have a description, if the probability of its top ranked candidate being a description is greater than $0.5$ (i.e.,a threshold of $0.5$), and no description otherwise.
Results for the \emph{All Identifiers} grouping are sensitive to the value of this threshold.  Although we only experimented with one value of this threshold, its possible that exploring different values for this threshold could result in improved performance of the NPR for the \emph{All Identifiers} grouping.  This threshold would also be a good parameter to adjust for potential applications and use cases with varying degrees of tolerance for false positive descriptions.
%
% Threshold...
The SOTA approach of \cite{Schubotz:2017} produces the best results for all identifier classes only containing identifiers with descriptions (i.e., the \emph{Explicit Descriptions}, \emph{Implicit Descriptions} and \emph{Any Descriptions} groupings).  For these identifier classes we always selected the top ranked description, regardless of whether the model predicts that no description is associated with the identifier.
As expected, due to the generative nature of the majority of the implicit examples, the extractive NPR approach performed significantly better on explicit examples (F1 score of $0.150$ for implicit examples versus $0.625$ for explicit examples).
\begin{table}
   \centering
    \setlength{\tabcolsep}{4pt}

\resizebox{\textwidth}{!}{
\begin{tabular}{lcccccccr}
\hline
\textbf{Identifier Class} & \textbf{Approach} & \textbf{Model} & \textbf{SF}  & \textbf{WV} & \textbf{CG} & \textbf{PCA} & \textbf{EM} & \textbf{F1}\\
\hline

\emph{All Identifiers}  & Variation & GBM           & 2          & No        & NNAN & No & 0.477 & 0.529\\
                & Variation & GBM           & 4           & Yes        & NNAN & No & 0.474 & 0.515\\
                & Variation & GBM           & 4           & No        & NC & No & 0.471 & 0.524\\
                &&&&\ldots \ldots \ldots&&&&\\
                & SOTA & SVM           & 1           & No        & NNAN & No & 0.326 & 0.402\\
\hline
\emph{Explicit Descriptions} & SOTA & SVM           & $\ast$           & $\ast$        & NNAN & No & 0.491 & 0.625\\
                & Variation  & VC          & 1               & Yes            & NNAN & No & 0.474 & 0.581\\
                & Variation  & SVM           & $\ast$           & $\ast$        & NC & No & 0.474 & 0.575\\
                
\hline
\emph{Implicit Descriptions} & SOTA & SVM           & $\ast$               & $\ast$        & NNAN & No & 0.000 & 0.150\\
                & Variation & SVM           & $\ast$             & $\ast$        & NC & No & 0.000 & 0.132\\
                & Variation & SVM           & 1                  & Yes        & NC & 50 & 0.000 & 0.128\\
\hline
\emph{Any Descriptions} & SOTA & SVM           & $\ast$             & $\ast$        & NNAN & No & 0.384 & 0.521\\
                 & Variation & SVM           & $\ast$           & $\ast$        & NC & No & 0.371 &	0.478\\
                 & Variation & VC            & 1                & Yes       & NC & No & 0.371 &	0.475\\

\hline
\end{tabular}}
  \caption{This table shows the best results achieved for the MIDR task on the MFQuAD dataset using different configurations of  the NPR approach. An asterisk ($\ast$) is used to indicate that all tested values for a configuration option produced the same results.  The following acronyms are used for the column names: \textbf{SF} for sampling factor - ratio of negative to positive examples, \textbf{WV} for word vectors, and \textbf{CG} (candidate generation).  Also NC represents noun chunk candidate generation.  The row with ellipses is used to indicate that some results are skipped in order to show the top SOTA result.
}
  \label{tab:npr-results}
\end{table}

In order to gain an insight into the contribution of individual features, we report the mean results across all configurations for when this feature is set and when it is unset.
The mean F1-scores for different options across all configurations in which they are enabled were 0.44 with word vectors, 0.39 without word vectors, 0.42 with NNAN candidate generation, 0.41 with noun chunk candidate generation, and 0.45, 0.42 and 0.38 for a negative sampling factor of 1, 2 and 4 respectively.
Thus, on average, across all configurations the best performing ones used word vectors, NNAN candidate generation and downsampled negative to positive samples to a 1:1 ratio.

The NPR approach clearly performs well for explicit examples, with the top configuration achieving an F1-Score of 0.625. This result is still significantly less than the state-of-the-art results across many other NLP tasks though (e.g., for the SQuAD dataset ~\cite{rajpurkar-etal-2018-know}), hinting that there is still room for significant improvement here.
The fact that the variations of the NPR approach that we evaluated on had little impact on the results achieved suggests that a more distinct strategy may be required, rather than a variation of the SOTA.
Further, the contrast in performance on explicit versus implicit examples, illustrates the importance of considering non-span or grounded generative examples in a realistic evaluation of the MIDR task.  Whilst, it is clear that the extractive NPR approach will achieve a zero score for this class of examples in the Exact Match metric, the top F1-Score of 0.15, is also significantly lower than for explicit examples.
Also, the top performing configurations for the grouping \emph{All Identifiers} differ from those for the grouping \emph{All Descriptions} showing the importance of considering examples without a description in MIDR datasets.

\section{Conclusions and Future Directions\label{sec:conclusions}}
% Mention something about needing to ground language models...
Solving the MIDR task is important to understanding the huge body of knowledge contained in mathematical equations in a scalable self-supervised manner. It is not only essential in understanding domain knowledge from math, physics, engineering, or commerce, but, also important to provide formal knowledge for large scale `world models' that will drive significant progress in AI towards the holy grail of common sense reasoning~\cite{url:lecun_interview_website}.
In this paper we presented significant progress to solve the MIDR task with the creation of the new MFQuAD dataset for validating new and existing approaches.
The MFQuAD dataset introduced new features and example types such as (grounded) generative examples and NULL examples that are not part of existing datasets.
We also performed a baseline evaluation with the SOTA approach on the MFQuAD dataset, and highlighted aspects of the MIDR task that require further work.
The lower performance of the SOTA approach on grounded generative examples reinforced the importance of including this category of example in a representative dataset for the MIDR task.
% Future Work.
We make MFQuAD publicly available and invite researchers to work on this problem. Our focus will be on approaches to extract non-span descriptions of identifiers as these are not well handled by existing methods.

\bibliographystyle{ieeetr}
\bibliography{custom,anthology}

\begin{appendices}
\section{Complete Implicit Example List\label{sec:appendix-implicit-egs}}
This appendix provides a complete list of all implicit examples in the MFQuAD dataset.  The format is title, identifier pairs.
\begin{itemize}
    \itemsep-0.2em
    \item 331 model, $I$
    \item 331 model, $\beta$
    \item Allan variance, $h_{-2}$
    \item Analytic capacity, $H$
    \item Anderson localization, $d$
    \item Backstepping, $\dot{u}_{x}$
    \item Backstepping, $e_{1}$
    \item Backstepping, $f_{x}$
    \item Backstepping, $g_{x}$
    \item Binomial theorem, $k$
    \item Binomial theorem, $x$
    \item Binomial theorem, $y$
    \item Block cipher mode of operation, $x$
    \item Borsuk's conjecture, $d$
    \item Clenshaw algorithm, $\alpha$
    \item Covariant derivative, $X_{1}$
    \item Covariant derivative, $X_{2}$
    \item Differentiation rules, $x$
    \item Dirac bracket, $p_{x}$
    \item Dirac bracket, $p_{y}$
    \item Dirac bracket, $x$
    \item Dirac bracket, $y$
    \item Divisor function, $i$
    \item Drucker–Prager yield criterion, $b_{3}$
    \item E (mathematical constant), $t$
    \item Ehrenfest model, $N$
    \item Ext functor, $A$
    \item Graphlets, $H$
    \item Kernel Fisher discriminant analysis, $\mathbf{m}_{1}$
    \item LTI system theory, $\omega$
    \item Legendre transformation, $y$
    \item List of RNA structure prediction software, $\mathrm{seqs}$
    \item Mayer–Vietoris sequence, $i$
    \item Mayer–Vietoris sequence, $j$
    \item Mayer–Vietoris sequence, $k$
    \item Mayer–Vietoris sequence, $l$
    \item Near sets, $x_{7}$
    \item Partition function (statistical mechanics), $E_{j}$
    \item Partition function (statistical mechanics), $j$
    \item Projections onto convex sets, $x$
    \item Projections onto convex sets, $y$
    \item Q-Vectors, $t$
    \item Quantum channel, $H_{B}$
    \item Sigma additivity, $A$
    \item Superconducting radio frequency, $R_{\text{s normal}}$
    \item Table of thermodynamic equations, $W$
    \item Table of thermodynamic equations, $p$
    \item Time-evolving block decimation, $N$
    \item Van der Waerden's theorem, $\varepsilon$
\end{itemize}

\section{Noun Phrase Ranking Approach Features\label{app:appendix-npr-features}}
This appendix lists all features used during the noun phrase ranking approach, organised by type.
%Boolean - Pattern Matching
The following pattern matching features are taken from \cite{Pagel2014} via \cite{Schubotz2016a}.
\begin{itemize}
\itemsep-0.2em
 \item [P01]  <candidate> <identifier> %matches\_definien\_id
 \item [P02]  <identifier> <candidate> %matches\_id\_definien
 \item [P03]  <identifier> (is|are) <candidate> %matches\_id\_isare
 \item [P04]  <identifier> (is|are) the <candidate> %matches\_id\_is\_the
 \item [P05]  <identifier> (is|are) denoted by <candidate> %matches\_x\_is\_denoted\_by
 \item [P06]  <identifier> (is|are) denoted by the <candidate> %matches\_x\_is\_denoted\_by\_the
 \item [P07]  let <identifier> be denoted by <candidate> %matches\_let\_x\_be\_denoted\_by
 \item [P08]  let <identifier> be denoted by the <candidate> %matches\_let\_x\_be\_denoted\_by\_the
 \item [P09]  <identifier> denote(s?) <candidate> %matches\_id\_denotes
 \item [P10]  <identifier> denote(s?) the <candidate> %matches\_id\_denotes\_the
\end{itemize}

The following additional pattern match type features, are a superset of features from \cite{Kristianto2014} via \cite{Schubotz2016a}, which previous authors refer to as basic features.

\begin{itemize}
    \itemsep-0.2em
    \item [B01] Colon between the identifier and the candidate? %matches\_colon
    \item [B02]  Just colon between the identifier and the candidate? %matches\_just\_colon
    \item [B03]  Comma between the identifier and the candidate? %matches\_comma
    \item [B04]  Just comma between the identifier and the candidate? %matches\_just\_comma
    \item [B05]  Other math between the identifier and the candidate? %matches\_math\_or\_id
    \item [B06]  Does the candidate contain another the identifier? %matches\_other\_id\_in\_definien
    \item [B07]  Is the candidate inside parentheses? %matches\_parentheses\_candidate
    \item [B08]  Is the identifier inside parentheses? %matches\_parentheses\_id
    \item [B09]  Does the identifier appear before the candidate? %matches\_id\_before\_definien
\end{itemize}

We include the following textual features, which are a superset of the text features from \cite{Kristianto2014} via \cite{Schubotz2016a}.  Text features are represented as a matrix of token counts and/or a matrix of unigram, bigram and trigram counts.
\begin{itemize}
\itemsep-0.2em
\item [T01] Two tokens from before and after the candidate. % get\_definien\_context\_text
\item [T02] POS tags of two tokens from before and after the candidate. % get\_definien\_context\_pos
\item [T03] Two tokens and their POS tags from before and after the candidate. %get\_definien\_context\_text\_pos
\item [T04] Unigram, bigram and trigram of feature T1.
\item [T05] Unigram, bigram and trigram of feature T2.
\item [T06] Unigram, bigram and trigram of feature T3.
\item [T07] Three tokens from before and after the identifier. % get\_id\_context\_text
\item [T08] POS tags of three tokens from before and after the identifier. % get\_id\_context\_pos
\item [T09] Three tokens and their POS tags from before and after the identifier. % get\_id\_context\_text\_pos
\item [T10] Unigram, bigram and trigram of feature T7.
\item [T11] Unigram, bigram and trigram of feature T8.
\item [T12] Unigram, bigram and trigram of feature T9.
\item [T13] Text between the identifier and the candidate. % get\_link\_text
\item [T14] POS tags of text between the identifier and the candidate. % get\_link\_pos
\item [T15] POS tags and text between the identifier and the candidate. % get\_link\_text\_pos
\item [T16] Unigram, bigram and trigram of feature T13.
\item [T17] Unigram, bigram and trigram of feature T14.
\item [T18] Unigram, bigram and trigram of feature T15.
\item [T19] Text of the first verb that appears between the identifier and the candidate % get\_first\_verb\_text,
\end{itemize}

We include the following dependency graph features, which are a superset of the dependency graph features from \cite{Kristianto2014} via \cite{Schubotz2016a}.  Dependency graphs are generated using spaCy's model `en\_core\_web\_lg' version 3.1.

\begin{itemize}
    \itemsep-0.2em
    \item [D01] Length of shortest path between the identifier and the candidate in the dependency graph. % get\_identifier\_definien\_distance\_dep\_graph
    \item [D02] Text of first three tokens on shortest path from the candidate to the identifier. % get\_definien\_id\_sp\_dep\_text
    \item [D03] POS tags of first three tokens on shortest path from the candidate to the identifier. % get\_definien\_id\_sp\_dep\_pos
    \item [D04] Text and POS tag of first three tokens on shortest path from the candidate to the identifier. % get\_definien\_id\_sp\_dep\_text\_pos
    \item [D05] Unigram, bigram and trigram of feature D02.
    \item [D06] Unigram, bigram and trigram of feature D03.
    \item [D07] Unigram, bigram and trigram of feature D04.
    \item [D08] Text of first three tokens on the shortest path from the identifier to the candidate. % get\_id\_definien\_sp\_dep\_text
    \item [D09] POS tags of first three tokens on shortest path from the identifier to candidate. % get\_id\_definien\_sp\_dep\_pos
    \item [D10] Text and POS tag of first three tokens on shortest path from the identifier to candidate. % get\_id\_definien\_sp\_dep\_text\_pos
    \item [D11] Unigram, bigram and trigram of feature D08.
    \item [D12] Unigram, bigram and trigram of feature D09.
    \item [D13] Unigram, bigram and trigram of feature D10.
    \item [D14] Direction of edge on shortest path incident to candidate.
    \item [D15] Direction of edge on the shortest path incident to the identifier.
\end{itemize}

We include the following Information Retrieval features which are from \cite{Pagel2014} via \cite{Schubotz2016a}.

\begin{itemize}
    \itemsep-0.2em 
    \item [I01] The token distance between the identifier and the candidate. % get\_identifier\_definien\_distance,
    \item [I02] Distance between the candidate and the first occurrence of the identifier. % get\_identifier\_definien\_sentence\_distance
    \item [I03] Relative term frequency of the candidate. %get\_relative\_term\_frequency\_definien
\end{itemize}

We also include the following word vector features that were not used by previous NPR approaches.  Embeddings are generated using spaCy's model `en\_core\_web\_lg' version 3.1.

\begin{itemize}
    \itemsep-0.2em
    \item [V01] Two tokens from before and after the candidate. %get\_definien\_context\_embedding,
    \item [V02] Three tokens from before and after the identifier. % get\_id\_context\_embedding
    \item [V03] The first verb between the candidate and the identifier. % get\_first\_verb\_embedding,
    \item [V04] The text between the candidate and the identifier. %get\_link\_embedding
\end{itemize}

%%%%%%%%%%%%%%%%%%%%%%%%%%%%%%%%%%%%%%%%%%%%%%%%%%%%%%%%%%%%%%%%%%%%%%%%%%%%%

\end{appendices}
\end{document}